\documentclass{book}

\usepackage[utf8]{inputenc}
\usepackage{pdfpages}
\usepackage{fancyhdr}
\usepackage{hyperref}

\usepackage{combelow}
\usepackage{newunicodechar}

\usepackage{ amstext, epstopdf, booktabs, verbatim, gensymb, geometry, lmodern}
\usepackage{tocloft}
\newlength{\standardchapnumwidth}
\AtBeginDocument{\setlength{\standardchapnumwidth}{\cftchapnumwidth}}

\newunicodechar{Ș}{\cb{S}}

\newcommand*\cpiType{Volume 3}
\newcommand*\Date{December 2022}
\newcommand*\Author{Federico Castagna\\ Jack Mumford\\ \cb{S}tefan Sarkadi \\Andreas Xydis}

\definecolor{myblue}{HTML}{154360}
\definecolor{emerald}{HTML}{3cb371}

\begin{document}

\newgeometry{margin = 0in}

\pagecolor{emerald}
\setlength{\fboxsep}{0pt}
\hfill \colorbox{myblue}{\makebox[3.22in][r]{\shortstack[r]{\vspace{3.3in}}}}%
\setlength{\fboxsep}{15pt}
\setlength{\fboxrule}{5pt}
\colorbox{white}{\makebox[\linewidth][c]{\includegraphics[width=1.3in]{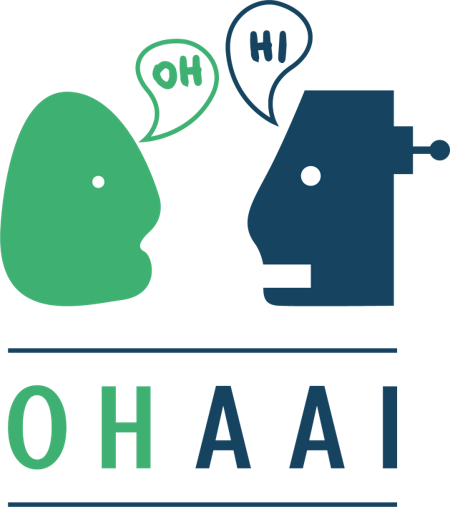}\hspace{0.35in} \shortstack[l]{\vspace{10pt}\fontsize{40}{40}\rmfamily\color{myblue} Online Handbook of \\
\vspace{10pt}
\fontsize{40}{40}\rmfamily\color{myblue} Argumentation for AI\\%
\fontsize{20}{20}\rmfamily\color{myblue} \cpiType}}}%
\setlength{\fboxsep}{0pt}
\vspace{-0.25pt}
\hfill \colorbox{myblue}{\hspace{.25in} \parbox{2.97in}{\vspace{2in} \color{white} \large{Edited by \\ \\ \Author  \\ \\  \Date \vspace{2.15in} \vfill}}}%
\restoregeometry

\nopagecolor

\thispagestyle{empty}
\pagenumbering{gobble}

\begin{center}
    \textbf{{\huge Preface}}
\end{center}

\hfill

This volume contains revised versions of the papers selected for the third volume of the Online Handbook of Argumentation for AI (OHAAI). Previously, formal theories of argument and argument interaction have been proposed and studied, and this has led to the more recent study of computational models of argument. Argumentation, as a field within artificial intelligence (AI), is highly relevant for researchers interested in symbolic representations of knowledge and defeasible reasoning. 
The purpose of this handbook is to provide an open access and curated anthology for the argumentation research community. OHAAI is designed to serve as a research hub to keep track of the latest and upcoming PhD-driven research on the theory and application of argumentation in all areas related to AI. The handbook’s goals are to:

\begin{enumerate}
    \item Encourage collaboration and knowledge discovery between members of the argumentation community.
    \item Provide a platform for PhD students to have their work published in a citable peer-reviewed venue.
    \item Present an indication of the scope and quality of PhD research involving argumentation for AI.
\end{enumerate}

The papers in this volume are those selected for inclusion in OHAAI Vol.3 following a back-and-forth peer-review process. The volume thus presents a strong representation of the contemporary state of the art research of argumentation in AI that has been strictly undertaken during PhD studies. Papers in this volume are listed alphabetically by author. We hope that you will enjoy reading this handbook.
\begin{flushright}
\noindent\begin{tabular}{r}
\makebox[1.3in]{}\\
\textit{Editors}\\
Federico Castagna\\
Jack Mumford\\
\cb{S}tefan Sarkadi\\
Andreas Xydis\\\\
\textbf{December 2022}
\end{tabular}
\end{flushright}


\pagenumbering{gobble}

\begin{center}
    \textbf{{\huge Acknowledgements}}
\end{center}

\hfill

\noindent
We thank the senior researchers in the area of Argumentation and Artificial Intelligence for their efforts in spreading the word about the OHAAI project with early-career researchers.

\hfill

\noindent
We are especially thankful to Sylwia Polberg and the COMMA 2022 organisers for collaborating with us at the conference summer school SSA 2022, during which OHAAI ran a student programme that provided the attending students the opportunity to present and discuss their PhD research amongst peers and academics in a friendly forum.

\hfill

\noindent
We are also grateful to ArXiv for their willingness to publish this handbook.

\hfill

\noindent
Our sincere gratitude to Costanza Hardouin for her fantastic work in designing the OHAAI logo.

\hfill

\noindent
We owe many thanks to Sanjay Modgil for helping to form the motivation for the handbook, and to Elizabeth Black and Simon Parsons for their advice and guidance that enabled the OHAAI project to come to fruition.

\hfill

\noindent
The success of the OHAAI project depends upon the quality feedback provided by our reviewers. We have been fortunate in securing a diligent and thoughtful program committee that produced reviews of a reliably high standard. Our thanks go to our PC: 

\noindent
Andreas Br{\"a}nnstr{\"o}m, Théo Duchatelle, Timotheus Kampik, Isabelle Kuhlmann, Lars Malmqvist, Mariela Morveli-Espinoza, Stipe Pand\v zi\' c, Christos Rodosthenous, Robin Schaefer, Kenneth Skiba, Luke Thorburn, Antonio Yuste-Ginel, and Heng Zheng.

\hfill

\noindent
Special thanks must go to the contributing authors: 

\noindent
Lars Bengel, Elfia Bezou-Vrakatseli, Lydia Bl{\"u}mel, Federico Castagna, Giulia D’Agostino, Daphne Odekerken, Minal Suresh Patil, Jordan Robinson, Hao Wu, and Andreas Xydis. Thank you for making the world of argumentation greater! 

\newgeometry{margin = 0.9in}

\pagenumbering{arabic}

\tableofcontents 
\thispagestyle{empty}
\clearpage




\pagestyle{fancy}
\addtocontents{toc}{\protect\renewcommand{\protect\cftchapleader}
     {\protect\cftdotfill{\cftsecdotsep}}}
\addtocontents{toc}{\setlength{\protect\cftchapnumwidth}{0pt}}

\refstepcounter{chapter}\label{1}
\addcontentsline{toc}{chapter}{On Serialisability for Argumentative Explanations \\ \textnormal{\textit{Lars Bengel}}}
\includepdf[pages=-,pagecommand={\thispagestyle{plain}}]{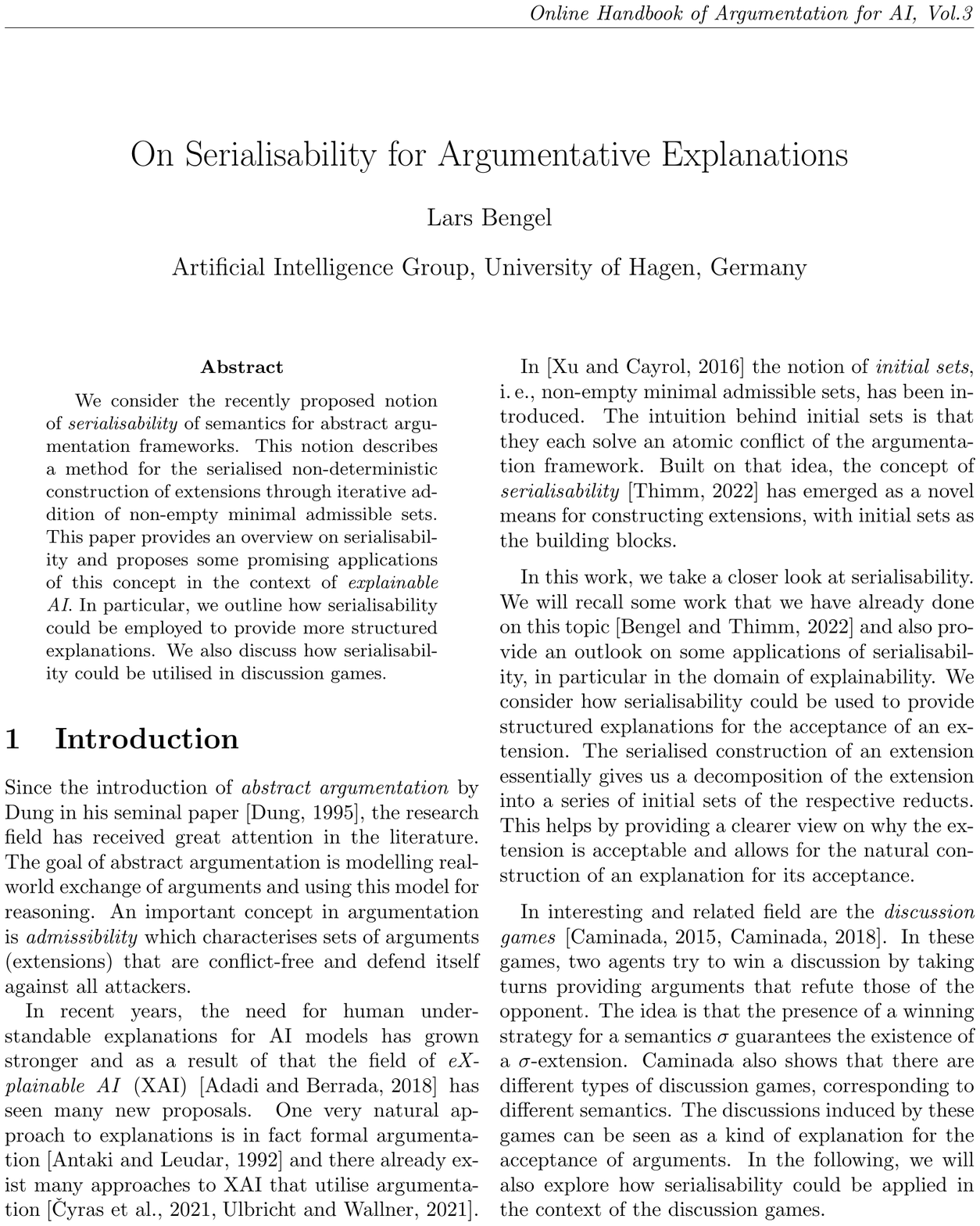}

\refstepcounter{chapter}\label{2}
\addcontentsline{toc}{chapter}{Debating Ethics: Using Argumentation to Support Dialogue \\
\textnormal{\textit{Elfia Bezou-Vrakatseli}}}
\includepdf[pages=-,pagecommand={\thispagestyle{plain}}]{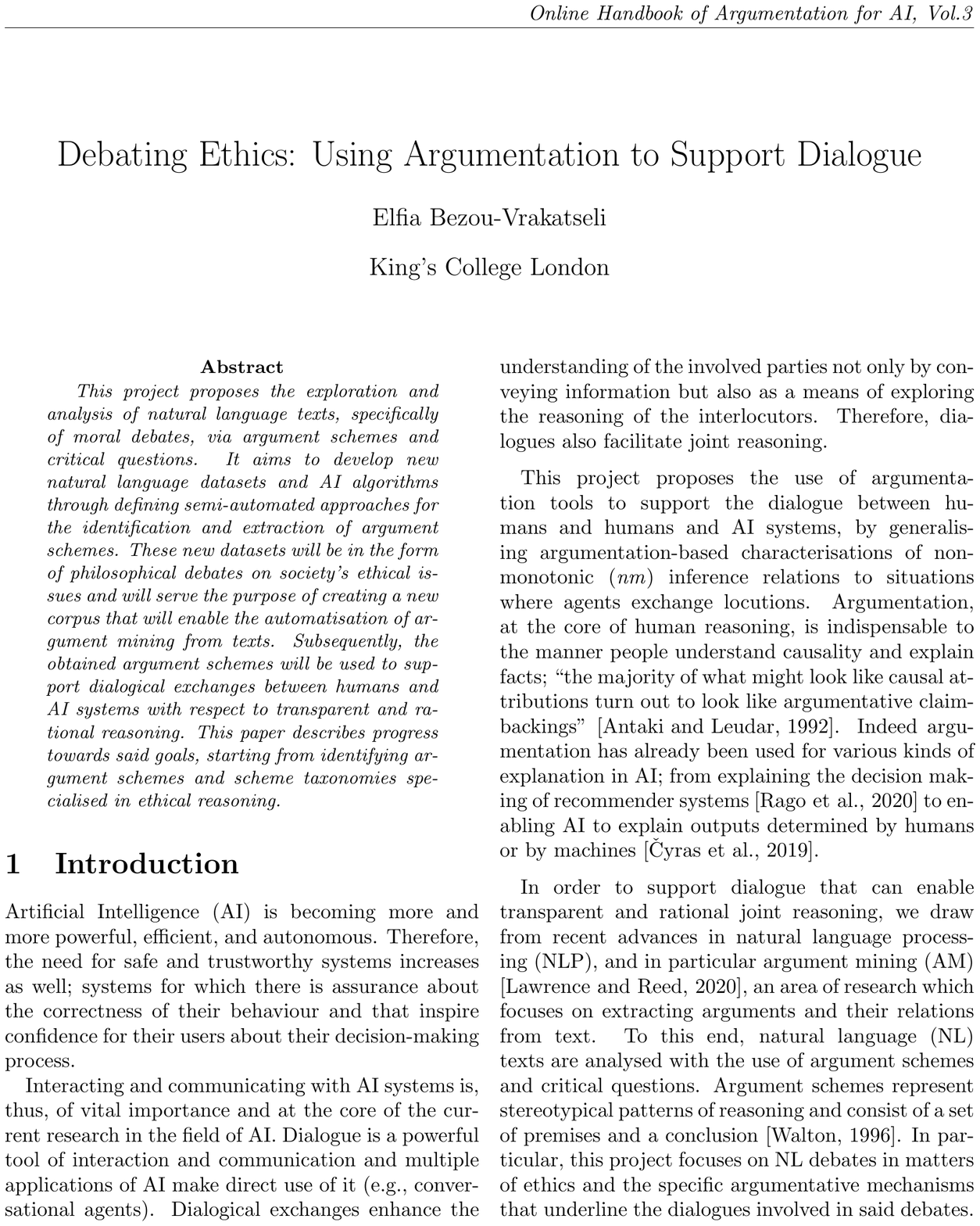}

\refstepcounter{chapter}\label{3}
\addcontentsline{toc}{chapter}{Characterization of Unresolvable Conflicts in Abstract Argumentation \\
\textnormal{\textit{Lydia Bl{\"u}mel}}}
\includepdf[pages=-,pagecommand={\thispagestyle{plain}}]{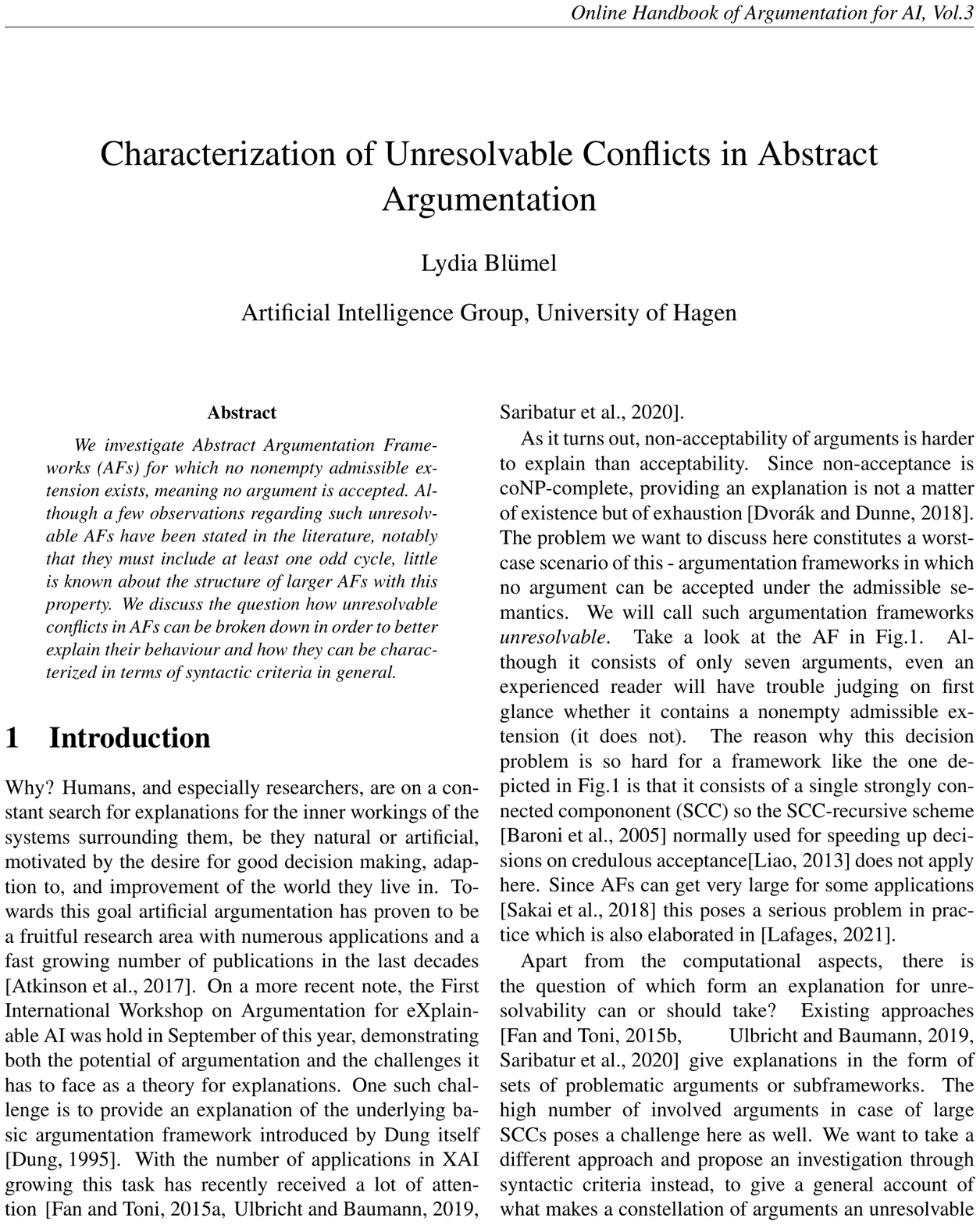}

\refstepcounter{chapter}\label{4}
\addcontentsline{toc}{chapter}{Towards a Fully-fledged Formal Protocol for the
Explanation-Question-Response Dialogue \\
\textnormal{\textit{Federico Castagna}}}
\includepdf[pages=-,pagecommand={\thispagestyle{plain}}]{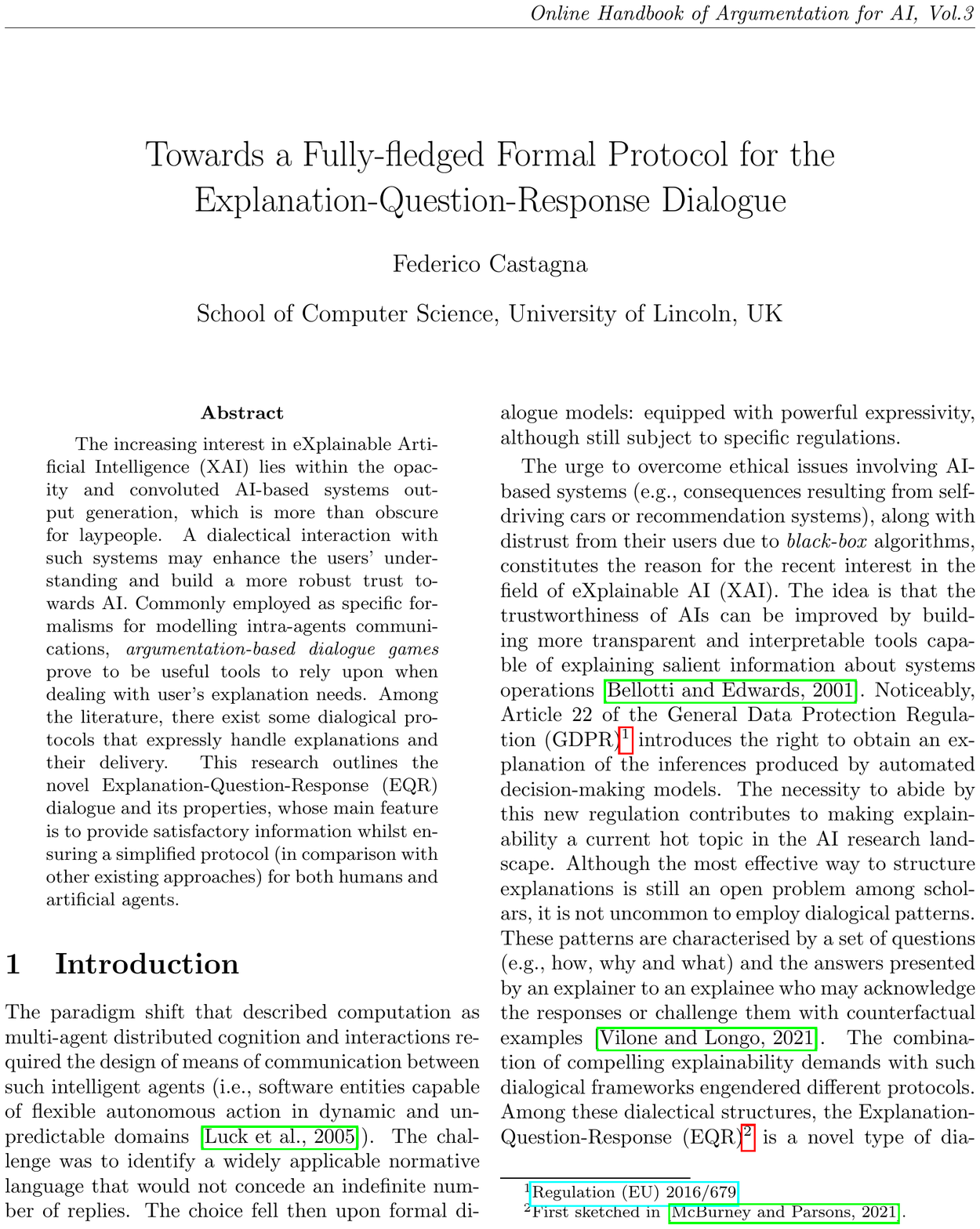}

\refstepcounter{chapter}\label{5}
\addcontentsline{toc}{chapter}{Argumentation without Opposition? \\
\textnormal{\textit{Giulia D’Agostino}}}
\includepdf[pages=-,pagecommand={\thispagestyle{plain}}]{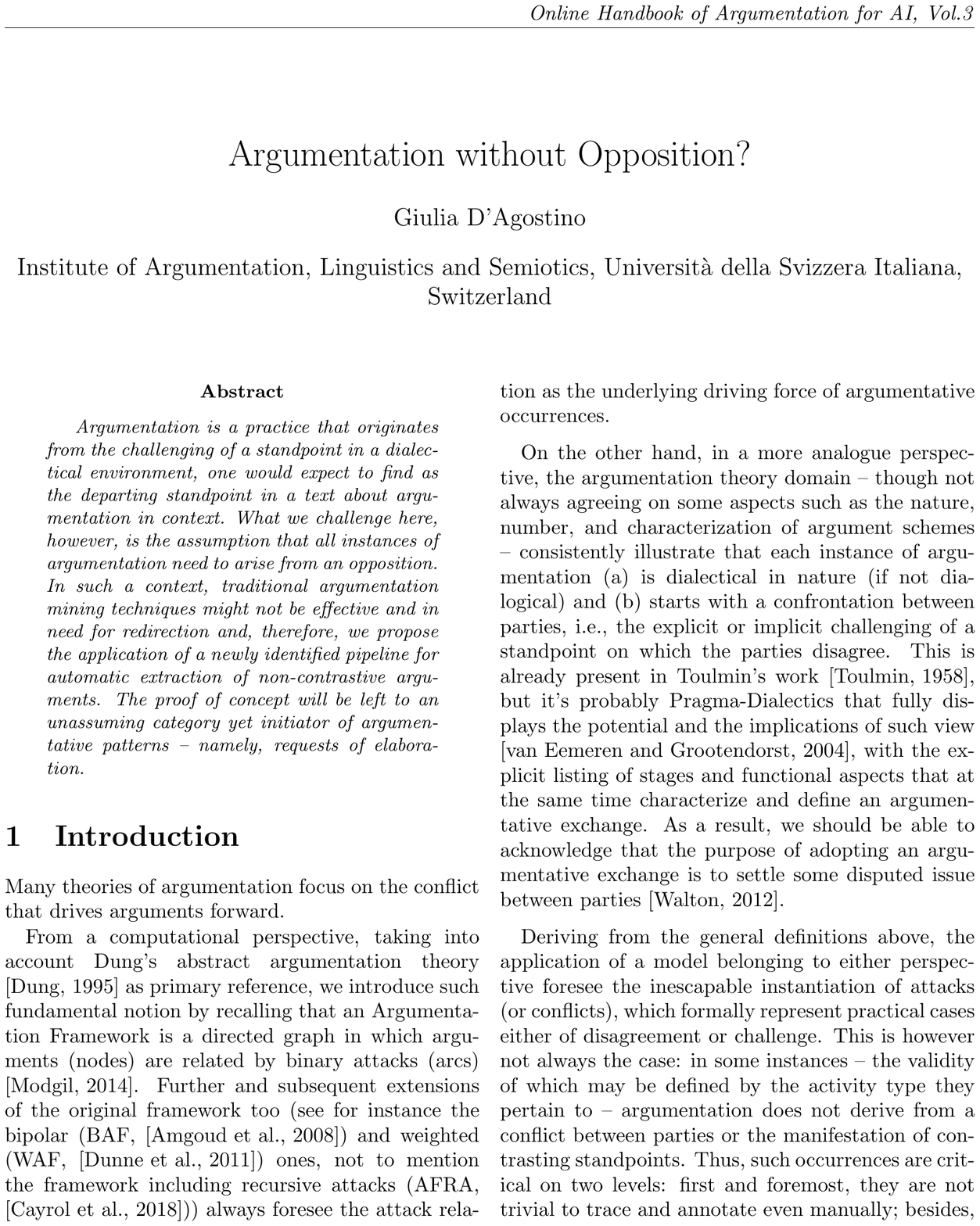}

\refstepcounter{chapter}\label{6}
\addcontentsline{toc}{chapter}{Justification, Stability and Relevance for Transparent and Efficient Human-in-the-Loop Decision Support \\
\textnormal{\textit{Daphne Odekerken}}}
\includepdf[pages=-,pagecommand={\thispagestyle{plain}}]{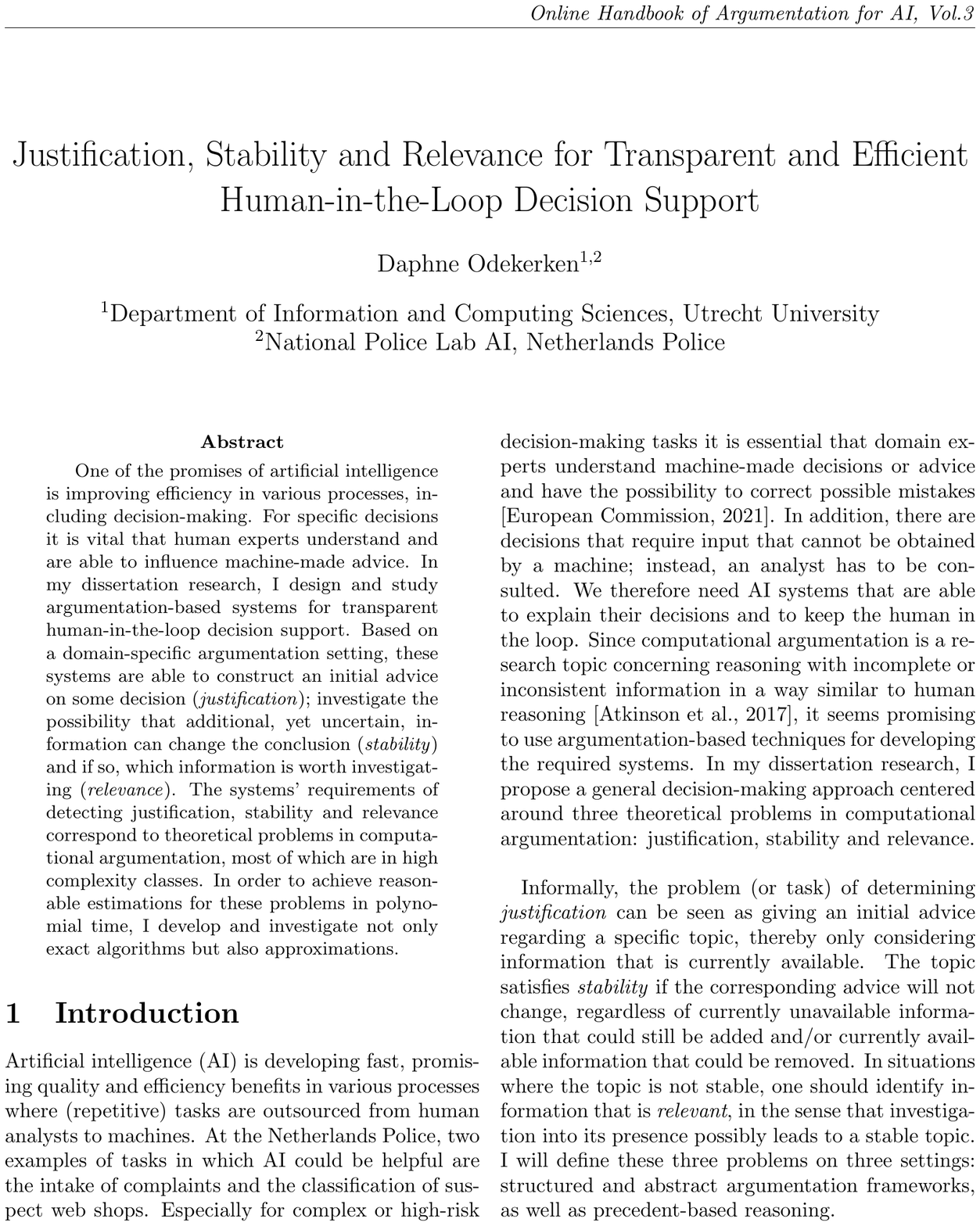}

\refstepcounter{chapter}\label{7}
\addcontentsline{toc}{chapter}{Towards Preserving Semantic Structure in Argumentative Multi-Agent via Abstract Interpretation \\
\textnormal{\textit{Minal Suresh Patil}}}
\includepdf[pages=-,pagecommand={\thispagestyle{plain}}]{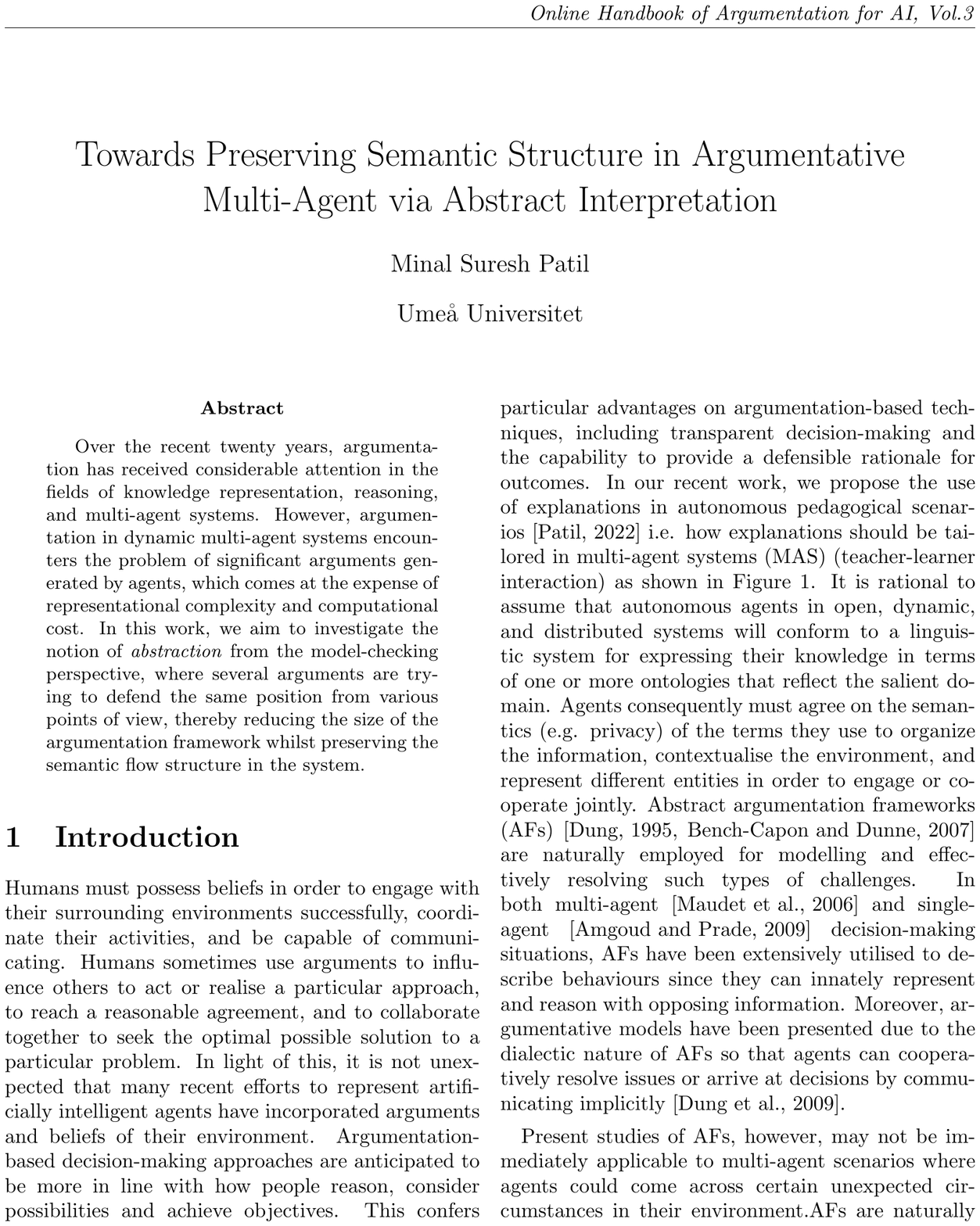}

\refstepcounter{chapter}\label{8}
\addcontentsline{toc}{chapter}{Distributed Hypothesis Generation and Evaluation \\
\textnormal{\textit{Jordan Robinson}}}
\includepdf[pages=-,pagecommand={\thispagestyle{plain}}]{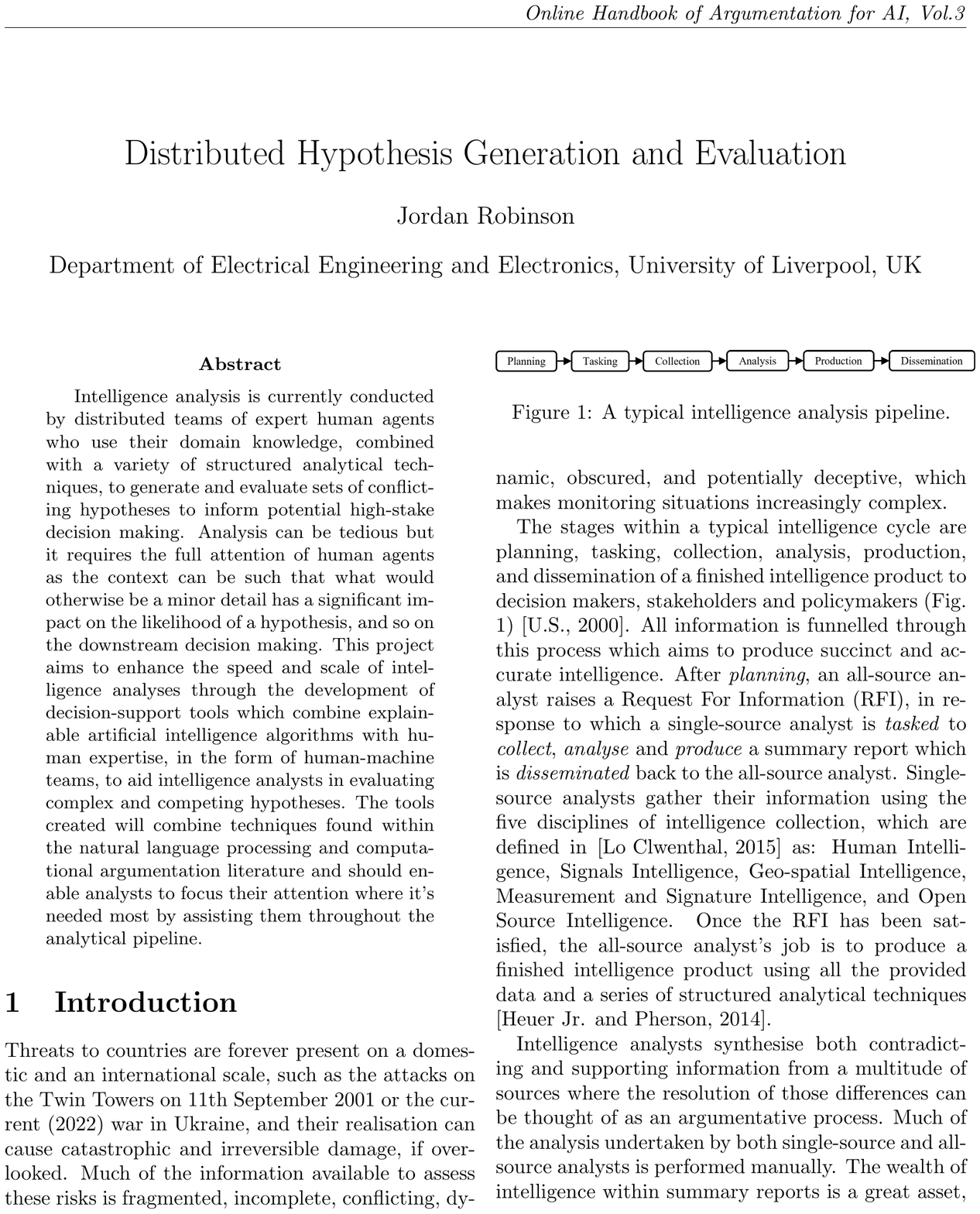}

\refstepcounter{chapter}\label{9}
\addcontentsline{toc}{chapter}{Exploring Internal Structures of an Argumentation System and Improving Reasoning Efficiency with Backward Searching Framework \\
\textnormal{\textit{Hao Wu}}}
\includepdf[pages=-,pagecommand={\thispagestyle{plain}}]{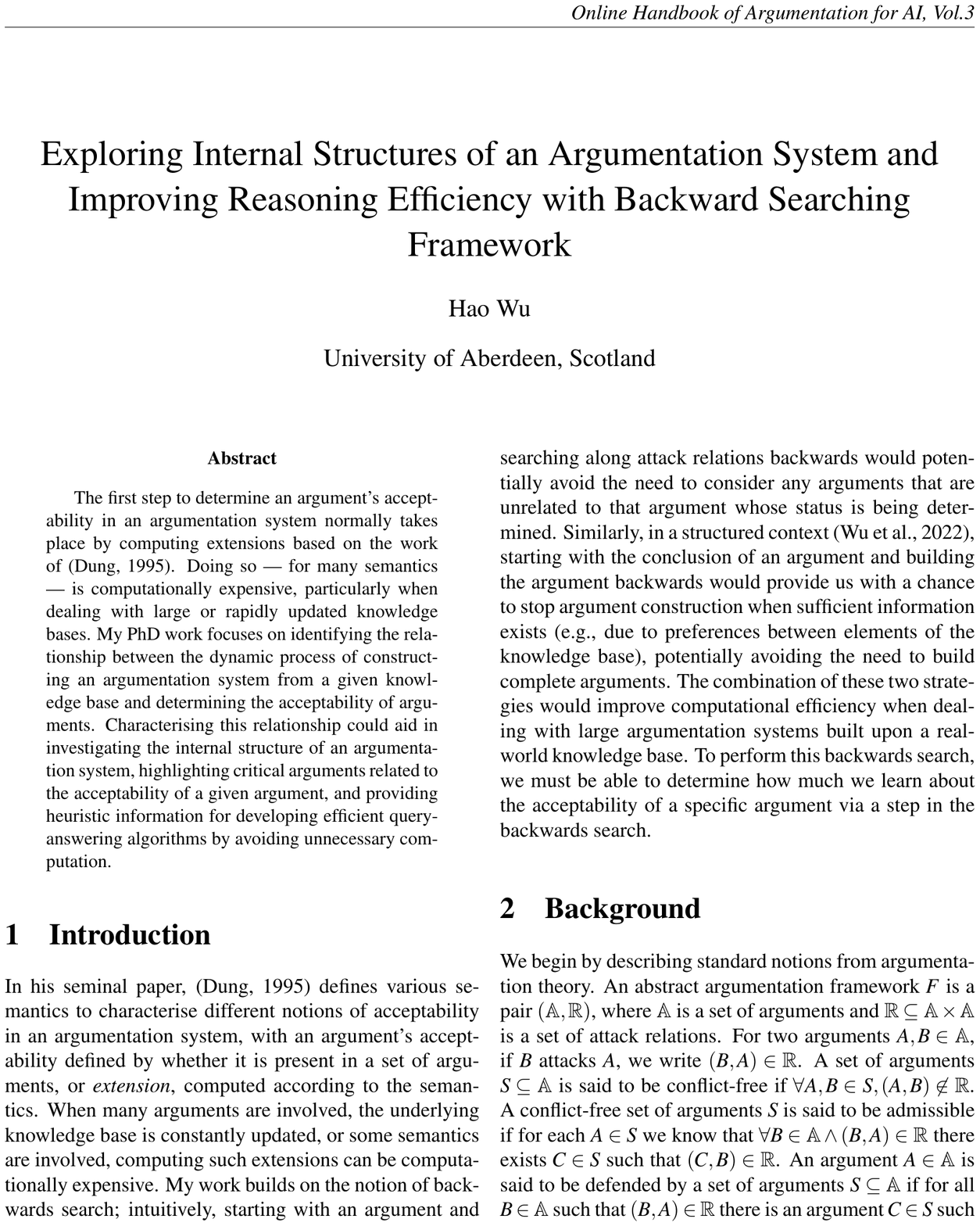}

\refstepcounter{chapter}\label{10}
\addcontentsline{toc}{chapter}{Discussing Soundness and Completeness for Dialogues that Account for Enthymemes \\
\textnormal{\textit{Andreas Xydis}}}
\includepdf[pages=-,pagecommand={\thispagestyle{plain}}]{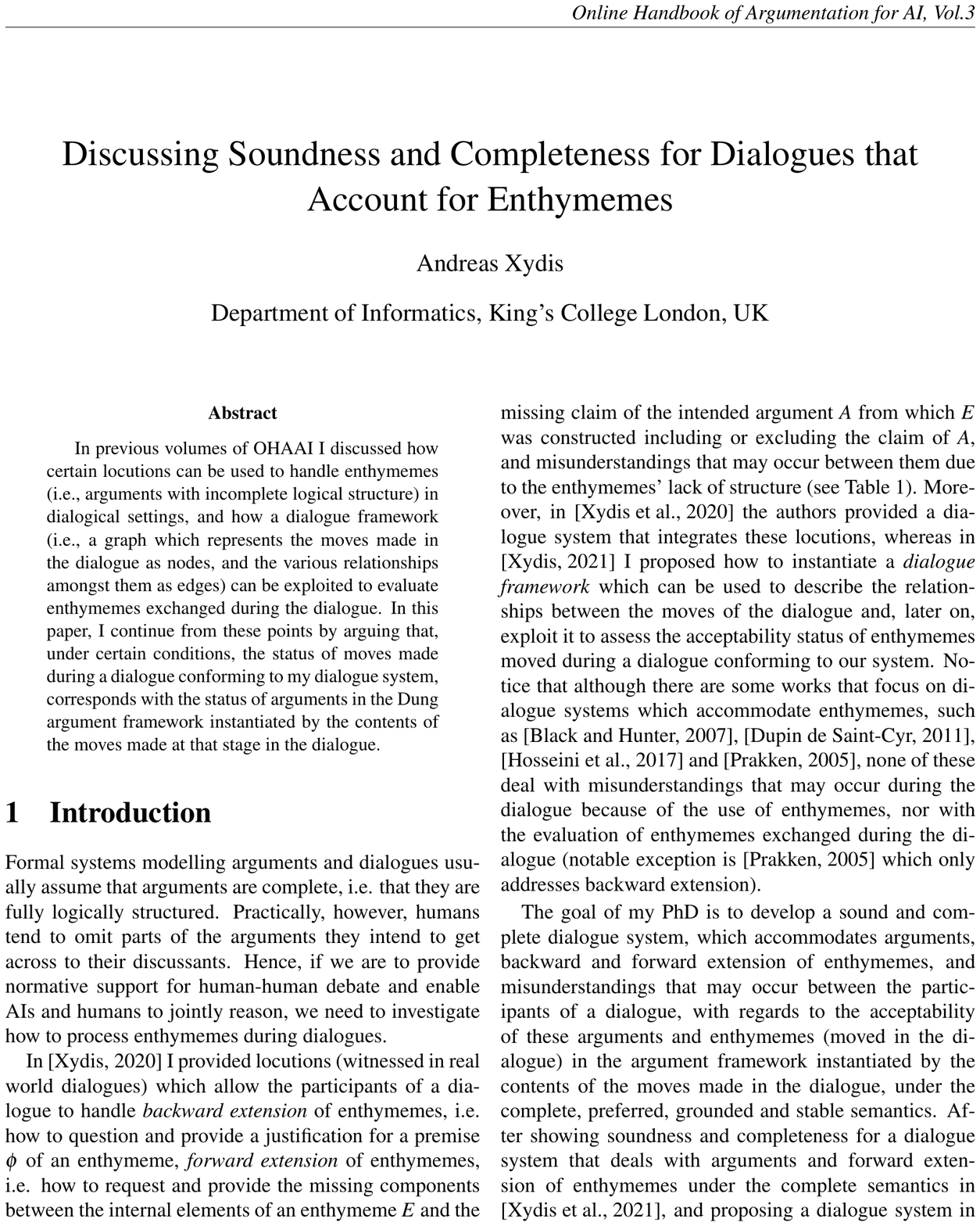}

\end{document}